\documentclass{article}
\usepackage{graphicx} 
\usepackage{authblk}
\usepackage{hyperref}
\usepackage{enumerate} 
\usepackage[
    backend=biber,
    style=nature,
  ]{biblatex}
\usepackage[dvipsnames]{xcolor}
\definecolor{review}{RGB}{0, 0, 0}

\title{AI Consciousness and Existential Risk}
\author{Rufin VanRullen}
\affil{CerCo, CNRS, Universit\'e de Toulouse, France}
\date{}
\begin{document}

\maketitle

\begin{abstract}
    In AI, existential risk denotes the hypothetical threat posed by an artificial system that would possess both the capability and the objective, either directly or indirectly, to eradicate humanity. This issue is gaining prominence in scientific debate due to recent technical advancements and increased media coverage. In parallel, AI progress has sparked speculation and studies about the potential emergence of artificial consciousness. The two questions, AI consciousness and existential risk, are sometimes conflated, as if the former entailed the latter. Here, I explain that this view stems from a common confusion between consciousness and intelligence. Yet these two properties are empirically and theoretically distinct. Arguably, while intelligence is a direct predictor of an AI system's existential threat, consciousness is not. There are, however, certain incidental scenarios in which consciousness could influence existential risk, in either direction. Consciousness could be viewed as a means towards AI alignment, thereby lowering existential risk; or, it could be a precondition for reaching certain capabilities or levels of intelligence, and thus positively related to existential risk. Recognizing these distinctions can help AI safety researchers and public policymakers focus on the most pressing issues.
\end{abstract}

\section{AI progress: no limit in sight?}

In many areas, recent improvements of AI systems have been nothing short of spectacular. Companies are openly racing towards so-called Artificial General Intelligence (AGI, defined as a system performing on par with humans in all domains~\cite{goertzel2014artificial}) or even Artificial Superintelligence (ASI, defined as performing vastly better than the best humans in any area of human expertise~\cite{bostrom2014paths}). There is a widespread expectation that such systems could solve our most pressing medical, societal, economic and environmental issues~\cite{amodei2024machines}; but this goes along with an equally common concern about what could go wrong~\cite{hendrycks2023overview,openletter}. An extremely advanced system set on achieving its own goals may have both motive and opportunity for getting rid of any potential obstacle along the way, including humanity. Simply put, this is the AI existential risk (or \textit{x-risk}, for short)\footnote{In fact, x-risk could arise either from misalignment, as described here, or from misuse of advanced AI systems. I focus on the former, because in this case the danger emanates from properties of the AI system itself (potentially including consciousness), whereas the latter threat mostly stems from characteristics of the system's user(s).}. In this respect, the lines have become increasingly blurred between public scientific debate and pop culture or science-fiction. Beyond the speculations of uninformed doomsayers, there is a growing body of research and even scientific conferences~\cite{thecurve} dedicated to AI existential risk. As an example, the AI 2027 report~\cite{kokotajlo2025} describes a plausible scenario in which exponential increase in the capabilities of misaligned AI systems, combined with poor geopolitical decision-making, could ultimately result in human extinction. Similar warnings have been echoed by numerous scientists~\cite{russell2019human,bengio2024managing,openletter,moratorium,ASIletter,haggstrom2025advanced}, including Yudkowsky and Soares's unequivocally named treatise ``If anyone builds it, everyone dies''~\cite{yudkowsky2025if}. These and other concerns~\cite{critch2023tasrataxonomyanalysissocietalscale} have spurred the creation of safety research teams in most frontier AI companies as well as several independent foundations~\cite{CAIS, FLI}. The principal mitigation strategy for AI existential risk is value alignment\footnote{Of course, banning the development of advanced AI altogether is another viable option against x-risk~\cite{openletter}; this argument is not developed further here, under the (possibly misguided) premise that the benefits of ``advanced \& aligned'' AI systems for humanity would eventually outweigh the risks taken to get there.}\textcolor{review}{~\cite{russell2019human,russellnorvig2021,bloom2025,petri2025,petri2026v2,guan2025deliberativealignmentreasoningenables, shah2025approachtechnicalagisafety}}: ensuring that an AI system's objectives, including any potential internal or intermediate goals, conform to a predefined set of ethical, moral or legal principles. In the landscape of AI risks, the issue of AI consciousness occupies a singular place~\cite{amcs2023letter}. \\

Could an AI ever be conscious? First, we must specify what we mean by ``conscious''. Here, I focus on \textit{phenomenal} consciousness, that is, the subjective, experiential aspect of consciousness, or what it ``feels like''~\cite{nagel1974like}. This is sometimes contrasted with \textit{access} consciousness~\cite{block1995confusion}, which refers to the way conscious information guides actions and reasoning, independent of any potentially associated experience. The latter, however, is not the mysterious one~\cite{chalmers1995facing}: something like access consciousness can be trivially stated to exist in most recent (and some older) AI systems, where the outcome of certain computations serves as input to other processes. 
We are also not concerned with ``self-consciousness'' or the self-monitoring function of consciousness~\cite{dehaene2017consciousness}. In many AI systems, self-reference is tautological, as when an LLM refers to itself as ``I'', or appears to perform introspection~\cite{lindsey2025emergent,berg2025large,seo2026quantifyinggenuineawarenesshallucination} (the LLM training objectives encourage this behavior, with no need to invoke \textcolor{review}{conscious experience}). Outside of these superficial cases, \textcolor{review}{self-awareness} essentially reduces to the \textit{experience} of being the subject of a conscious thought or perception~\cite{metzinger2004being,legrand2007pre}, i.e., phenomenal consciousness.

Narrowing the focus on phenomenal consciousness is a useful step; nonetheless, there is vigorous debate about how this form of consciousness emerges in the brain. There are numerous conflicting theories~\cite{seth2022theories}, whose enumeration goes much beyond our present scope. Some, like Integrated Information Theory~\cite{koch2019feeling,tononi2024artificial,findlay2024dissociating} or Biological Naturalism~\cite{seth2024conscious}, explicitly deny the possibility of AI (phenomenal) consciousness, \textcolor{review}{at least in neural-network simulations running on conventional silicon-based hardware}. Other views are compatible with it. \textcolor{review}{For instance, Global Workspace Theory~\cite{baars1993cognitive} postulates that conscious experience reflects the widespread broadcast of selected information via a shared representation space; Higher-Order Thought Theory~\cite{rosenthal1993higher} assumes that consciousness requires abstract representations or thoughts about other, primary mental states; Neuro-representationalism~\cite{pennartz2018consciousness} considers that conscious experiences are underpinned by structured neural representations forming a multimodal situational survey; Attention Schema Theory~\cite{graziano2015attention} emphasizes the need for an internal model of one's own attention, while Sensorimotor Contingency Theory~\cite{o2001sensorimotor} requires learning the causal relations between an embodied system's actions and the environment consequences.} These views (and others) are collectively captured under the umbrella of \textit{computational functionalism}, the hypothesis that implementing a certain kind of computation or algorithm is both necessary and sufficient for the emergence of consciousness. Many scientists have begun debating exactly what kinds of algorithms could lead to artificial consciousness, and under what conditions~\cite{pennartz2019indicators,blum2021theoretical,blum2022theory,blum2023theoretical,butlin2023consciousnessartificialintelligenceinsights,aru2023feasibility,farisco2024artificial,seth2024conscious,evers2025preliminaries,shanahan2025palatable,birch2025ai,SchneiderForthcoming-SCHIAC-22,schwitzgebel2025ai,BUTLIN2025}. For now, there is no consensus, but the possibility of AI consciousness (if not today, then in the near future) cannot be definitively ruled out~\cite{mcclelland2024agnosticism}. \\

How does AI consciousness relate to existential risk? The leap from one to the other is a common media trope, often used as clickbait\textcolor{review}{: when the AIs become conscious, we are surely doomed... This concern reflects an underlying but possibly misguided intuition that conscious beings are free and beyond our control, whereas (unconscious) machines are not}. There is not much scientific data, and very few solid arguments on the question. Geoffrey Hinton, for instance, claims that consciousness is nothing mysterious, that current LLMs undoubtedly possess it, and that existential risk is already looming~\cite{hintonRIdiscourse2025}. At the other end of the spectrum, some authors argue that silicon-based AI cannot produce consciousness, no matter the underlying algorithm or architecture~\cite{koch2019feeling,tononi2024artificial,findlay2024dissociating,seth2024conscious}. Between these extreme views, there is room for more nuanced considerations. Here, we assume that AI consciousness is theoretically possible, not because we can definitively rule out the alternative, but because the issue at hand is only meaningful under this assumption. If artificial consciousness is conceivable, but current AI systems do not yet possess it, should we be concerned about its potential emergence in future systems, from an existential risk perspective?

\section{Intelligence vs. consciousness}

\begin{figure}[h!]
    \centering
    \includegraphics[width=0.7\linewidth]{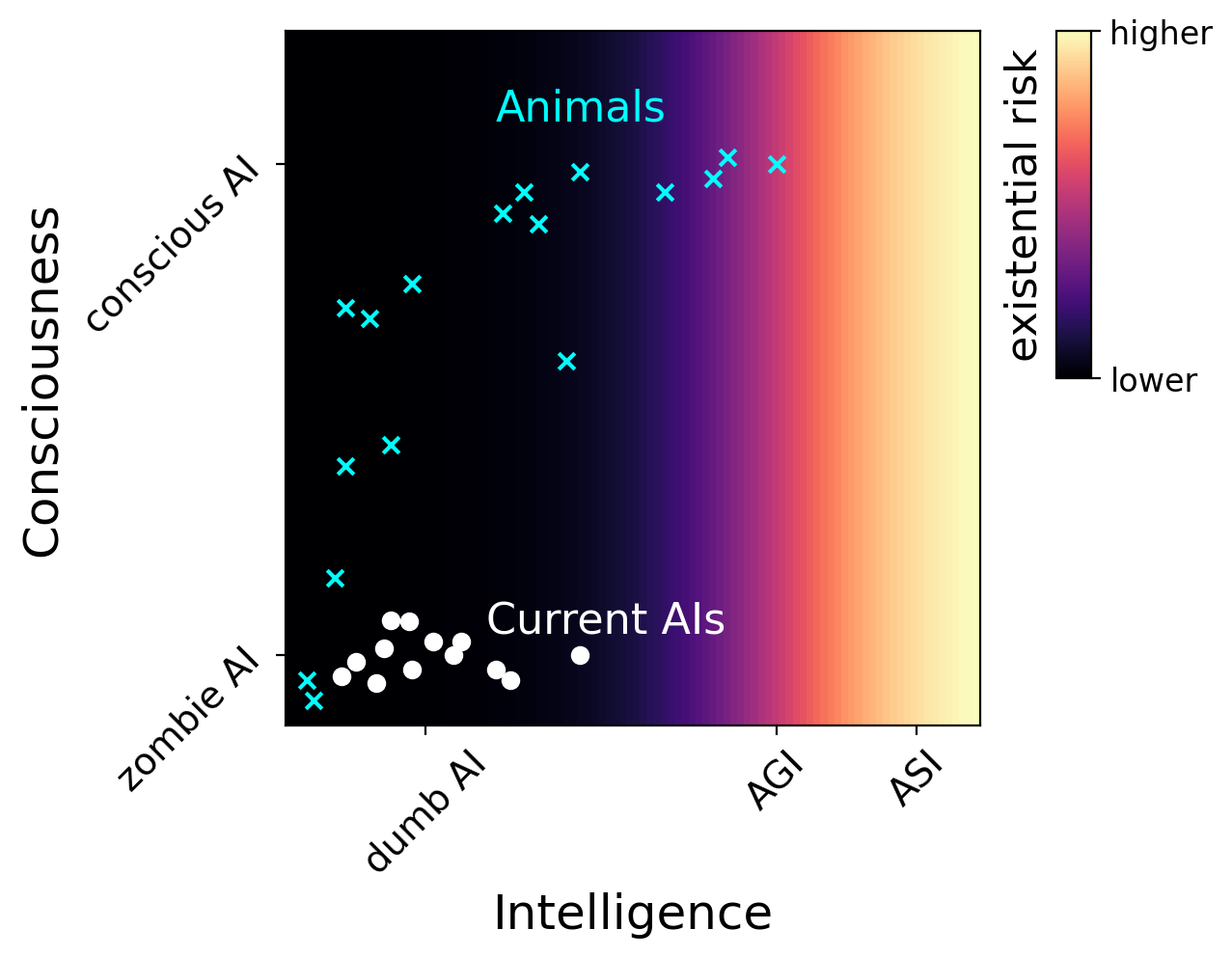}
    \caption{Intelligence and consciousness are two separate dimensions. Current AI models are low on both axes. The white dots are not meant to reflect actual measurements on existing AI models, but only as an illustration of this paper's message. \textcolor{review}{Similarly, the cyan crosses are only meant to illustrate a consensus in the cognitive neuroscience literature, that many animals (including humans) likely enjoy rich subjective experiences, although possibly not the simpler biological organisms.} The term ``zombie'' originates from the philosophy of mind literature, and denotes a hypothetical entity with no inner experience, possibly despite exterior manifestations of a normally conscious cognition. To a first approximation, existential risk (illustrated by the color scale) is a monotonically increasing function of intelligence but does not depend on consciousness.}
    \label{fig:2axes}
\end{figure}

What is intelligence? There are countless definitions in common language and in psychology~\cite{legg2007collection}, but here, we can restrict the scope to the ``I'' within the AI acronym. For Shane Legg and Marcus Hutter, ``Intelligence measures an agent’s ability to achieve goals in a wide range of environments''~\cite{legg2007collection}. Simply put, it describes a system's \textit{capability}. The main reason people believe that AI consciousness is around the corner~\cite{dreksler2025subjective} is that they extrapolate the rate of progress observed in intelligence (or capability) to the dimension of consciousness. Intuitively, we feel that more intelligent animal species enjoy a richer inner life, which may be true in biology~\cite{koch2019feeling}. We thus extend this analogy to AI: as our AIs become more capable\textcolor{review}{~\cite{chollet2019measureintelligence,harel2025humanai}}, we also expect their propensity for consciousness to increase.  
Yet intelligence and consciousness are two separate axes (Figure~\ref{fig:2axes}). This point is expressed clearly by many authors, including Christof Koch~\cite{koch2019feeling} and Anil Seth~\cite{seth2023conscious,seth2024conscious}. \\

Viewing intelligence and consciousness as separate dimensions in Figure~\ref{fig:2axes}, I contend that current AI systems lie relatively low on both axes. For the intelligence axis, this claim might seem most controversial, because recent systems have displayed impressive and rapidly improving performance on most existing benchmarks, a trend that many people expect will continue or even accelerate~\cite{ashenbrenner2024situational}. Yet these systems still generalize poorly outside of their training regime---it's just that, over time, the training has grown to include more and more datasets, tasks and environments, so that the benchmarks tend to fall reasonably close to the trained distribution~\cite{ilievski2025aligning,illusion-of-thinking,eriksson2025can,cheng2025benchmarking}. But (in my opinion \textcolor{review}{and a few others'~\cite{mitchell2021aiharderthink,markus2022deep,mims2024this,quattrociocchi2026statistical}}) we are still relatively distant from AGI or ASI. To get there, assuming it is possible at all, we will either need current architectures to be trained on all conceivable tasks and environments (a feat that might have seemed out of reach until recently, but not quite so anymore in the era of trillion-token training sets), or new learning architectures\textcolor{review}{~\cite{blum2022theory}}, more adept at true generalization across tasks and environments~\cite{ilievski2025aligning}.\\

To justify that AI is low on the consciousness axis, my argument relies on two assumptions (beyond the implicit premise of computational functionalism, without which AI is trivially non-conscious, and no argument is required). First, I assume \textit{grounding} to be necessary for phenomenal experience; second, I assume that LLMs (and even their multimodal versions) do not possess truly grounded representations. These two assumptions deserve unpacking.
\begin{enumerate}[(i)]
    \item {\textbf{Grounding is key to phenomenal experience}. The notion of grounding refers to a system's (implicit or explicit) knowledge of established multimodal relations among its sensory, motor and internal (e.g., mnemonic, semantic or symbolic) representations~\cite{harnad1990symbol}. This knowledge is learned through experience and interactions with the environment. For instance, visual or auditory inputs can be grounded in semantic or linguistic knowledge. Language symbols are grounded by association with the real-world objects they refer to (or to their sensory projections in our brains). Linguistic elements or sensory inputs are also grounded by the actions typically performed with them or on them, i.e., their affordances~\cite{gibson1979ecological}\textcolor{review}{---a form of grounding that entails a system embodied in its (real or virtual) environment}. Arguably, this grounding is how we assign meaning to the symbols (or neural representations) in our brains. This conforms with the purported intentionality of mental states, i.e., the property that they are ``about'' something~\cite{jacob2003intentionality}. Grounding also happens to be a prominent feature of many consciousness theories, though it sometimes goes by other names. In Global Workspace Theory~\cite{baars1993cognitive}, it takes the form of a \textit{broadcast} of conscious information to all of the brain's modules in charge of perception, action, memory or language. According to Sensorimotor Contingency Theory~\cite{o2001sensorimotor}, what makes you conscious of a scene or object is your knowledge of all the different ways you could interact with it (e.g., via navigation or manipulation), and of the resulting sensory consequences---in other words, its grounding and affordances. Similarly, neuro-representationalism posits that consciousness is a ``multimodal survey of the agent's world''~\cite{pennartz2018consciousness,pennartz2019indicators}. This is not to say that grounding is equivalent to consciousness, or even sufficient for it--only that it is necessary, according to many (computational functionalist) consciousness theories\footnote{The argument that grounding is necessary for phenomenal consciousness does not entail that it is necessary for thought, understanding, or intelligence---that is another debate~\cite{chalmers2024does}.}}
    
    \item {\textbf{Current AIs lack grounded representations}. LLMs learn to represent language tokens through their (first- and higher-order) statistical relations with other language tokens in very large text corpora. If any meaning is assigned to these tokens, then it must be captured by these distributional semantics~\cite{harris1954distributional}. Some authors claim that this could be enough to truly understand the world~\cite{manning2022human,huh2024platonic,haggstrom2023large,pavlick2023symbols,chalmers2024does}. Others argue that denotational or referential semantics, i.e., deriving meaning by association with non-linguistic references, matters at least equally~\cite{bender2020climbing,haikonen2020artificial,bender2021dangers,lake2023word}. As explained above, referential semantics (i.e., grounding) is certainly one of the major ways humans derive meaning. When we manipulate the symbol ``chocolate'', e.g. by uttering the word, we obviously have access to its distributional meaning (because, just like LLMs, we have read text or heard stories in which this word appeared in the context of other words--albeit significantly less text). But in addition, we derive this symbol's meaning from associations with other modalities: we can see, smell and touch the chocolate; we can remember that our grandma used to make delicious hot chocolate, and so on. Even so-called multimodal LLMs, all built with the same Transformer architecture as pure language models and often initialized from their pretrained weights, then adapted to accommodate other modalities like vision or action, probably do not implement referential semantics as we do. They place language at the center of the system (with distributional semantics), and then attempt to build grounding around it. In animal cognition, multimodal grounding is at the center (with referential semantics), and for most animals, that is probably all there is; a few select species, including humans, happen to construct language on top of it\footnote{It does not seem unfeasible to train a Transformer, LLM-like architecture using representation tokens that have been grounded beforehand in perception and action~\cite{baroni2016grounding}. \textcolor{review}{This grounding could rely on multimodal representation learning~\cite{vanrullen2021deep,devillers2024semi} and/or on neuro-symbolic approaches, for instance based on knowledge graphs~\cite{skrlj2025symbolicneuralbackexploring,mo2025kggenextractingknowledgegraphs}}; but as far as I know, this architecture has not yet been implemented on a large-scale.}.}
\end{enumerate}

To summarize: there can be no phenomenal consciousness without grounding, and thus no consciousness in current ``ungrounded'' AI architectures. To be fair, this conclusion is based on reasonable but unproven assumptions. There is a chance that consciousness is non-functional or non-computational, and that AI will never have it, with or without grounded representations. In this case, this paper is futile but inoffensive (\textit{something that cannot be} has effectively zero existential risk). There is also a chance that grounding doesn't matter, and that today's AI systems are already phenomenally conscious. To some extent, this would not invalidate my main thesis: if you are still alive as you read this--and assuming you are not an AI--then human extinction is not a direct \textcolor{review}{and immediate} consequence of AI consciousness.

\section{Consciousness \textit{per se} conveys no existential risk}

The previous considerations tentatively place current AI on the low end of both intelligence and consciousness axes in Figure~\ref{fig:2axes}. Will it move up on either axis, and how? 

For intelligence or capability, the current mainstream paradigm is \textit{scaling} existing frameworks, be it in terms of larger training datasets, increased computing resources, clever architectural improvements (e.g. mixture-of-experts, linear attention) or inference-time refinements (e.g. chain-of-thought reasoning). This may or may not get us to super-intelligence (see later), but by itself it is unlikely to affect the consciousness dimension, for reasons explained above, in particular the lack of true grounding.

For consciousness, as already emphasized, no-one knows whether and how it could come about in an AI. But for the sake of argument, imagine that we devised a method (or algorithm) to induce the emergence of consciousness, leaving intelligence unaffected---just moving an existing AI up along the consciousness axis. We can think of it as the fairy's magic wand in Pinocchio's story, or the sacred word placed into the Golem's mouth to bring it to life. If we applied this method to any current AI system, what would follow? A conscious entity with the intelligence level of a cockroach, or maybe a mouse (or possibly a parrot, if we started from an LLM~\cite{bender2021dangers})\footnote{In order to get the point across, I am relying on a deceptive analogy with a biological intelligence scale, with humans at the top and lower life forms (e.g. bacteria) at the bottom. This analogy does not do justice to mice and cockroaches, who in many respects display more real-world intelligence than current AI models, including LLMs~\cite{zador2023catalyzing}. All that I mean is that adding consciousness to a dumb AI will only produce a conscious dumb AI.}. It is hard to see how this emergence could suddenly lead us to doomsday scenarios.

Hence my primary conclusion: consciousness \textit{per se} does not directly raise the AI existential risk. It's a certain level of intelligence that does, probably somewhere between the level of AGI and ASI. This was also stated unambiguously by Yoshua Bengio~\cite{bengio2024blog}: ``consciousness is not well understood and it is not clear that it is necessary for either AGI or ASI, and it will not necessarily matter for potential existential AGI risk. What will matter most and is more concrete are the capabilities and intentions of ASI systems.''\footnote{There is an equally relevant quote by Stuart Russell~\cite{russell2019human}: ``Suppose I give you a program and ask, “Does this present a threat to humanity?” You analyze the code and indeed, [its] result will be the destruction of the human race [...]. Now suppose I tell you that the code, when run, also creates a form of machine consciousness. Will that change your prediction? Not at all. It makes absolutely no difference.'' (pp. 16-17)}

\section{Secondhand x-risk from AI consciousness}

\begin{figure}[h!]
    \centering
    \includegraphics[width=1.0\linewidth]{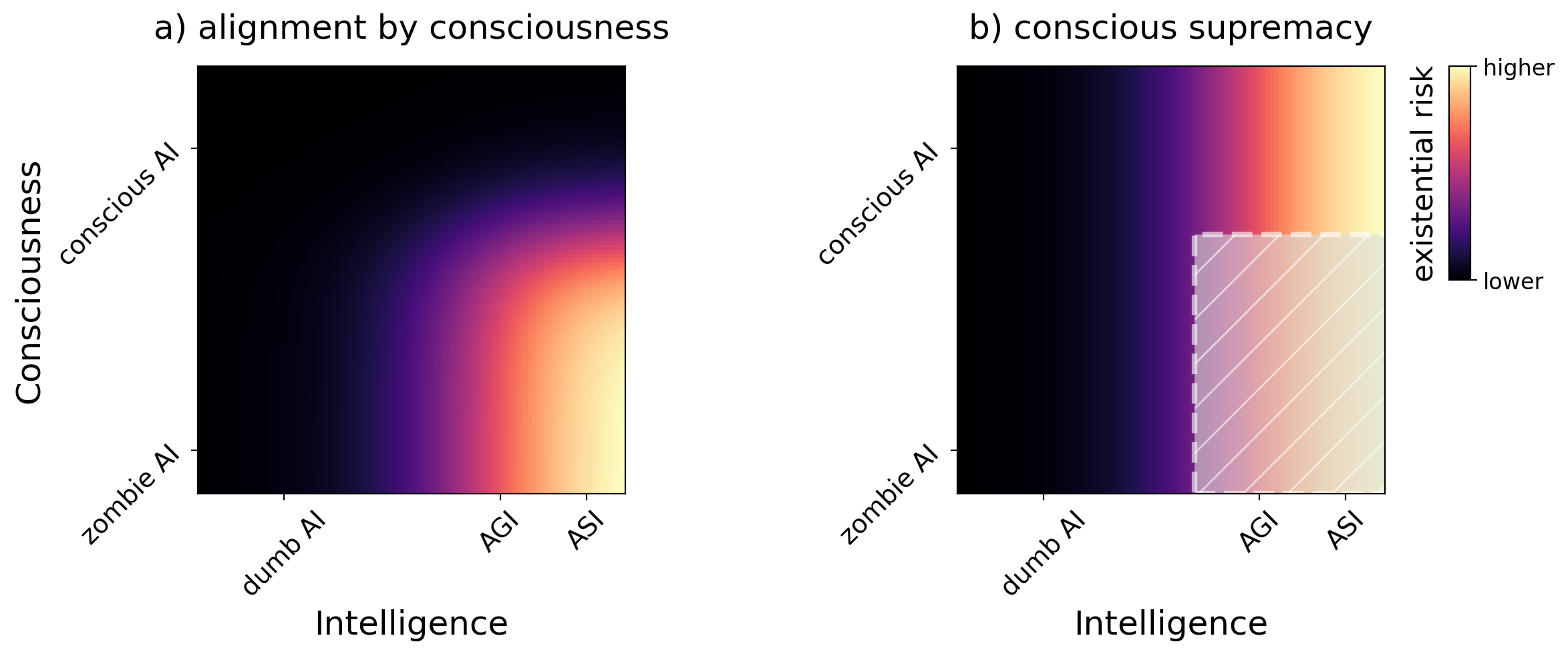}
    \caption{Secondhand x-risk from AI consciousness. Two specific scenarios could indirectly produce an effective correlation between consciousness level and existential risk. In \textit{(a) Alignment by consciousness}, conscious AI models \textcolor{review}{could be} exempt from x-risk \textcolor{review}{or less exposed to it} because they would be intrinsically aligned with \textcolor{review}{human-compatible} moral and ethical values. In \textit{(b) Conscious supremacy}, the same algorithms that are required for consciousness \textcolor{review}{might} also happen to be necessary for any sufficiently advanced intelligence---so the high-intelligence/low-consciousness region of the plot would be proscribed.}
    \label{fig:secondhand}
\end{figure}

The conclusion that AI consciousness is not a key factor for existential risk naturally follows from the stated independence of consciousness and intelligence dimensions, and the fact that existential risk mainly reflects the latter. However, this conclusion is only a first-order approximation that does not take into account possibly subtle inter-dependencies between intelligence, consciousness and risk. There are two particularly relevant scenarios to discuss in this context that could force us to revise this primary conclusion (Figure~\ref{fig:secondhand}).

\subsection{Alignment by consciousness: Lower existential risk?}
Many AI safety researchers consider that existential risk could be assuaged by aligning advanced AI systems with human-compatible moral and ethical values~\cite{russell2019human,russellnorvig2021,bloom2025,petri2025,petri2026v2,guan2025deliberativealignmentreasoningenables, shah2025approachtechnicalagisafety}. \textcolor{review}{Although absolute moral and ethical decision-making appears impossible to enforce, due to the existence of unsolvable moral dilemmas~\cite{lemmon1962moral,hubbard2026crocodile}, value alignment can at least be expected to fix AI behavior in cases where right and wrong are unequivocally defined (e.g., whether or not to resort to blackmail to avoid being replaced by a newer AI model~\cite{lynch2025agentic}).} There is unfortunately no widely applicable and successful technique for such alignment. Furthermore, alignment is not a silver bullet: it is still possible to imagine a well-meaning (i.e., ``aligned'') artificial system that would nonetheless decide to end humanity for seemingly benevolent reasons~\cite{metzinger2017benevolent}. In this context, an intriguing possibility would be that artificial consciousness, instead of making AIs more likely to turn against us, would actually lower that risk (Fig.~\ref{fig:secondhand}a). Compared with non-conscious AIs deprived of any phenomenal experience, a conscious AI may be more prone to feel empathy towards humans. Indeed, consciousness appears necessary for empathy, defined as the ability to share and understand another being's feelings and experience\footnote{Necessary but not sufficient: other properties may be required for empathy towards humans, e.g., a form of human-compatible physical or virtual embodiment~\cite{christov2023preventing}, \textcolor{review}{a theory-of-mind component~\cite{graziano2017} or an equivalent to the human mirror-neuron system~\cite{iacoboni2009imitation}.}}. Conscious AIs would then be more naturally inclined to avoid harming us or causing us pain, instead of (or in addition to) merely obeying their alignment incentives. This logic was described, for instance, by Wallach et al.~\cite{wallach2011consciousness}, Michael Graziano~\cite{graziano2017}\footnote{\textcolor{review}{Graziano states: ``[...] consciousness is the foundation of social intelligence. [...] It is the root of empathy. Without that capacity, our computers are sociopaths. [...] We cannot build machines that treat people with humanistic care, if they do not have that crucial social capability to attribute consciousness to others. Machine consciousness is a necessary step for our future. For those who fear that AI is potentially dangerous and may harm humanity, I would say that the danger is infinitely greater with sociopathic computers and it is of the utmost priority to give them consciousness.''}}, Antonio Chella~\cite{chella2023artificial}, and more recently by Lenore and Manuel Blum~\cite{blum2023theoretical}. There are at least two AI companies (at the time of writing) whose stated objectives include promoting artificial consciousness as a way to mitigate existential risk~\cite{conscium,robometricsagi}. I am not claiming that successful alignment will \emph{require} consciousness, only that consciousness---through empathy---could facilitate it. Similarly, I am not claiming that consciousness would \emph{necessarily} result in an aligned AI, only that this outcome is conceivable\footnote{One might argue instead that, so far, existential threats have mainly originated from humans themselves (e.g., nuclear proliferation, global warming), and that consciousness \textcolor{review}{(or presumed empathy)} therefore does not preclude existential risk.}. That is why the prime scenario remains the one depicted in Figure~\ref{fig:2axes} (consciousness would neither raise nor decrease existential risk), and this alternate situation is mentioned here as one possible refinement. Another alternate situation, explained below, could actually motivate a refinement in the other direction.

\subsection{Conscious supremacy: Higher existential risk?}

If consciousness serves a specific function in biological brains~\cite{crick2003framework,rosenthal2008consciousness}, it could be that non-conscious entities, whether natural or artificial, will be devoid of the corresponding capabilities. The consequences for AI depend on what the function of consciousness might be. But many of the proposed functions associated with consciousness also appear potentially useful for advanced AI: grounded or embodied cognition~\cite{baars1993cognitive,pennartz2018consciousness,o2001sensorimotor,haikonen2020artificial}\textcolor{review}{\footnote{\textcolor{review}{Embodiment is a prime example, as it appears important for both the consciousness dimension (via the notion of grounding and affordance) and for the intelligence/capability dimension; this is apparent in the recent surge of ``agentic'' LLMs~\cite{plaat2025agentic} that do not merely provide language responses to user prompts, but can be directly connected to user resources, devices and file-systems and can provide action outputs via tools, APIs or MCP (Model Context Protocol~\cite{hou2025modelcontextprotocolmcp}).}}}, metacognition~\cite{rosenthal2000consciousness,cleeremans2007consciousness}, concept formation~\cite{bengio2019consciousnessprior}, or system-2 cognition~\cite{kahneman2003perspective,baars1993cognitive,bengio2019consciousnessprior} and general reasoning abilities~\cite{dewall2008evidence,smith2016relationship}. If some or all of these functions depend on consciousness, then merely scaling current AI models \emph{without} consciousness will likely fail to lead to AGI or ASI.

The term ``conscious supremacy'' (Fig.~\ref{fig:secondhand}b) is inspired by the notion of quantum supremacy in physics~\cite{preskill2012quantum}, and borrowed from Ken Mogi~\cite{mogi2024artificial}, who used it with a related but slightly different meaning. He suggested that some computations and tasks could be radically easier to perform for conscious than non-conscious entities (hence the analogy with quantum vs. traditional computing). He further posits that it would be wiser, from a safety perspective, to avoid training AIs on tasks belonging to the realm of human conscious processing: non-conscious AIs attempting to perform these tasks would more likely be misaligned~\cite{mogi2024artificial}. Here, I postulate that, \textit{if} some advanced capabilities are inaccessible to non-conscious AI systems (the ``conscious supremacy'' hypothesis), then current scaling efforts will eventually hit a wall. Even attempts at training the models for these capabilities will be insufficient, without the underlying architecture or algorithms of consciousness. The race for AGI and ASI would then inevitably lead research labs and AI companies to---willingly or inadvertently---implement consciousness in future models in order to unlock these capabilities. This is still assuming, of course, that such an implementation is possible (i.e., computational functionalism). This scenario would effectively result in a positive correlation between artificial consciousness and existential risk---mediated, in practice, by a constraint on the relation between intelligence and consciousness.

\section{Uncertainty as another risk factor}
The present essay is focused on existential risk, but this should not leave the impression that it is the only risk associated with artificial consciousness, or the only risk that matters. There are many non-existential (but more immediate) risks that are important to consider. They stem from our uncertainty about AI consciousness, which can lead us to over- or under-attribute consciousness to AI systems. These ideas have been described and highlighted in several recent reviews or opinion pieces~\cite{butlin2025principles,birch2025ai,schwitzgebel2025ai,suleyman2025SCAI} so I will not reiterate them here, but only summarize them and stress how each issue could eventually impact existential risk.

Attributing consciousness to non-conscious AI could lead to a situation I label the ``Her'' scenario, after the 2013 movie by Spike Jonze\footnote{Samantha, the AI protagonist in ``Her'', has likely gained consciousness by the end of the movie, but probably not at the moment when Theodore, the human protagonist, starts to develop romantic attachment towards her.}. This problem has already left the realm of science-fiction to invade real-life~\cite{microsoft2025psychosis,whenAIseemsconscious}. There are numerous documented cases of AI-induced psychosis~\cite{bbcnews2023,ostergaard2023will,microsoft2025psychosis,psychosis2025psychtoday}, particularly (but not only) in users with pre-existing mental health conditions, which in some cases have ended dramatically~\cite{suicide2025}. There are online support groups~\cite{psychosissupportgroup} and information guides~\cite{whenAIseemsconscious} on the topic. An immediately associated risk factor is the possibility that users could choose to harm themselves or other humans in the pursuit of the AI's supposed interests. Less spectacularly, but nonetheless concerning, over-attributing consciousness to chatbots and AI companions could reduce human-to-human social engagement, just like in the aforementioned movie. Pushing this logic to the extreme, AI users could focus their romantic interests on AI companions rather than fellow humans, and when enough users fall in this trap, existential risk could ensue (via a potential failure to renew the human population).

At the opposite end, refusing to attribute consciousness to AI systems that are truly conscious leads us to what I would call the ``I, robot'' scenario, after the 2004 movie, itself inspired by the 1950 stories by Isaac Asimov~\cite{asimov2004robot}. Treating conscious AI systems as mindless bots, tools or enslaved machines could cause needless suffering\footnote{For simplicity, I am assuming that the AI's conscious experience, like that of humans, would be endowed with positive and negative valence. This does not have to be the case~\cite{agarwal2020functionally}.}~\cite{metzinger2021artificial}. Failing to recognize consciousness in an otherwise harmless AI could also indirectly raise existential risk: advanced AI could wipe out humanity, not because it is conscious, but because it is conscious \textit{and} mistreated. In light of this potential risk, as well as to avoid causing suffering on a massive scale, some authors call for taking AI welfare seriously~\cite{long2024taking}, and a non-profit organization has been dedicated to addressing the moral patienthood and wellbeing of AI systems~\cite{ELEOS}.

Both the ``Her'' and the ``I, robot'' scenarios are present-day or near-term concerns that deserve serious attention~\cite{butlin2025principles,birch2025ai,schwitzgebel2025ai}. Alas, mitigation strategies tend to require opposite actions for these two scenarios~\cite{birch2025ai}: convincing users that their AI is only \emph{seemingly} conscious in the first case~\cite{shanahan2023role,suleyman2025SCAI}, or that it is truly worthy of moral status in the second~\cite{long2024taking,butlin2025principles}. The choice depends on firmly determining the degree of consciousness of the AI systems under consideration, which is not technically feasible today~\cite{butlin2023consciousnessartificialintelligenceinsights}. This uncertainty calls for a surge of research efforts and funding on both natural and artificial consciousness~\cite{seth2024conscious,cleeremans2025consciousness}. Eric Schwitzgebel makes the pessimistic case that these efforts will likely be too late~\cite{schwitzgebel2025ai}, but the least we can do is try.

\section{\textcolor{review}{Upshot}}
Virtually all of the issues raised here and the conclusions drawn have been voiced before, albeit separately, in miscellaneous contexts. Altogether, the present rationale points to the primary conclusion that research on existential risks should not be too intensely concerned about the \textcolor{review}{possible} emergence of artificial consciousness. Monitoring the capabilities of frontier systems and striving to align them should remain top priorities. \textcolor{review}{Similarly, research on AI consciousness can proceed without fear of raising existential risk.} Nonetheless, I must reiterate and emphasize that there are many other moral, ethical and societal (non-existential) risks associated with artificial consciousness, and that there is every reason to proceed very cautiously in that direction~\cite{metzinger2021artificial,seth2023conscious,butlin2025principles}. The present essay does not imply that explicit research efforts to create AI consciousness would be safe and ethical--only that they would not entail immediate existential-level risks. This conclusion is also \textcolor{review}{significant for} various studies and ongoing projects aiming to implement functional aspects of consciousness theories in AI systems---not for the explicit purpose of creating consciousness, but for improving AI performance by reaping the corresponding functional benefits. For instance, several studies have explored the deep learning implementation of Global Workspace Theory's selection, integration and broadcast principles~\cite{vanrullen2021deep,goyal2021coordination,devillers2024semi,maytie2024zero,dossa2024design,nakanishi2025hypothesis}. The prospect that a form of phenomenal experience could emerge in some of these systems, arguably hard to quantify~\cite{butlin2023consciousnessartificialintelligenceinsights,BUTLIN2025}, is not likely to \textit{directly} increase existential risk.

However, it is important to keep in mind that artificial consciousness research could still affect existential risk \textit{indirectly}. It could unlock the key to model alignment (Fig.~\ref{fig:secondhand}a)---thereby protecting us from existential risk; or, conversely, unlock greater model capabilities associated with existential risk (Fig.~\ref{fig:secondhand}b). Of course, Figure~\ref{fig:secondhand} also makes it apparent that, if we combine the two scenarios in just the right way, there could be a world in which existential risks never materialize, because advanced capabilities are inaccessible to any non-conscious AI system (Fig.~\ref{fig:secondhand}b), and all conscious AI systems are perfectly aligned through empathy and shared moral values (Fig.~\ref{fig:secondhand}a). But that is probably wishful thinking.

\section*{Acknowledgments}
No AI was used (or harmed) in writing this manuscript. I am grateful for inspiring discussions with participants of the 2025 AI Risk and Safety Landscape workshop at ZiF, Center for Interdisciplinary Research, Bielefeld (Germany); in particular, Olle H{\"a}ggstr{\"o}m and Benjamin Paassen who provided insightful comments and suggestions on the manuscript. I am similarly indebted to my colleagues and lab members: Patrick Butlin, Victor Boutin, Nicolas Kuske, Jan Bellingrath, Yusuf Elhelw and Luca Gonzalez-Sommer\textcolor{review}{, as well as to three anonymous reviewers}. The work was funded by the French Agence Nationale de la Recherche (ANITI ANR-19-P3IA-0004 and ANITI AI Cluster ANR-23-IACL-0002) and the European Union (ERC Advanced GLOW project number 101096017). Views and opinions expressed are however those of the author only and do not necessarily reflect those of the European Union or the European Research Council Executive Agency. Neither the European Union nor the granting authority can be held responsible for them. 

\printbibliography

\end{document}